\documentclass[11pt,a4paper]{article}
\usepackage[hyperref]{acl2021}
\usepackage{times}
\usepackage{latexsym}

\usepackage{graphicx}
\usepackage{float}
\usepackage{amsmath}

\usepackage{caption}
\usepackage{subcaption}
\usepackage[T1]{fontenc}
\usepackage[utf8]{inputenc}
\usepackage{microtype}

\usepackage{microtype}

\aclfinalcopy 

\title{Can Sequence-to-Sequence Models Crack Substitution Ciphers?}

\author{Nada Aldarrab \and Jonathan May  \\
University of Southern California \\
Information Sciences Institute \\
\texttt{\{aldarrab,jonmay\}@isi.edu}}

\date{}

\begin{document}
\maketitle
\begin{abstract}
Decipherment of historical ciphers is a challenging problem. The language of the target plaintext might be unknown, and ciphertext can have a lot of noise. State-of-the-art decipherment methods use beam search and a neural language model to score candidate plaintext hypotheses for a given cipher, assuming the plaintext language is known. We propose an end-to-end multilingual model for solving simple substitution ciphers. We test our model on synthetic and real historical ciphers and show that our proposed method can decipher text without explicit language identification while still being robust to noise. 
\end{abstract}

\section{Introduction}

Libraries and archives have many enciphered documents from the early modern period. Example documents include encrypted letters, diplomatic correspondences, and books from secret societies (Figure \ref{fig:historical-ciphers}). Previous work has made historical cipher collections available for researchers \citep{pettersson-megyesi-2019-matching, megyesi-2020}. Decipherment of classical ciphers is an essential step to reveal the contents of those historical documents. 

In this work, we focus on solving 1:1 substitution ciphers. Current state-of-the-art methods use beam search and a neural language model to score candidate plaintext hypotheses for a given cipher \cite{kambhatla-2018}. However, this approach assumes that the target plaintext language is known. Other work that both identifies language and deciphers relies on a brute-force guess-and-check strategy \cite{knight-etal-2006-unsupervised,hauer-2016}. We ask: Can we build an end-to-end model that deciphers directly without relying on a separate language ID step?

The contributions of our work are:
\begin{itemize}
    \item We propose an end-to-end multilingual decipherment model that can solve 1:1 substitution ciphers without explicit plaintext language identification, which we demonstrate on ciphers of 14 different languages.
    \item We conduct extensive testing of the proposed method in different realistic decipherment conditions; different cipher lengths, no-space ciphers, and ciphers with noise, and demonstrate that our model is robust to these conditions.
    \item We apply our model on synthetic ciphers as well as on the Borg cipher, a real historical cipher.\footnote{\url{https://cl.lingfil.uu.se/~bea/borg/}} We show that our multilingual model can crack the Borg cipher using the first 256 characters of the cipher.
\end{itemize}

\section{The Decipherment Problem}

Decipherment conditions vary from one cipher to another. For example, some cleartext might be found along with the encrypted text, which gives a hint to the plaintext language of the cipher. In other cases, called  \textbf{known-plaintext attacks}, some decoded material is found, which can be exploited to crack the rest of the encoded script. However, in a \textbf{ciphertext-only attack}, the focus of this paper, the cryptanalyst only has access to the ciphertext. This means that the encipherment method, the plaintext language, and the key are all unknown.

In this paper, we focus on solving 1:1 substitution ciphers. We follow \citet{nuhn-2013} and \citet{kambhatla-2018} and use machine translation notation to formulate our problem. We denote the ciphertext as $ f_1^N = f_1 \dots f_j \dots f_N $ and the plaintext as $ e_1^M = e_1 \dots e_i \dots e_M $.\footnote{Unless there is noise or space restoration, $N=M$; see Sections~\ref{sec:noise} and \ref{sec:nos-ciphers}.}

In a \textbf{1:1 substitution cipher}, plaintext is encrypted into a ciphertext by replacing each plaintext character with a unique substitute according to a substitution table called the \textbf{key}. For example: the plaintext word ``doors'' would be enciphered to ``KFFML'' using the substitution table:

\begin{table}[H]
\centering
\begin{tabular}{c|c}
 \textbf{Cipher} & \textbf{Plain} \\
\hline
K & d \\
F & o \\
M & r \\
L & s \\
\hline
\end{tabular}
\end{table}

The decipherment goal is to recover the plaintext given the ciphertext.

\section{Decipherment Model}

Inspired by character-level neural machine translation (NMT), we view decipherment as a sequence-to-sequence translation task. The motivation behind using a sequence-to-sequence model is:
\begin{itemize}
    \item The model can be trained on multilingual data \cite{gao-2020}, making it potentially possible to obtain end-to-end multilingual decipherment without relying on a separate language ID step.
    \item Due to transcription challenges of historical ciphers (Section~\ref{sec:noise}), ciphertext could be noisy. We would like the model to have the ability to recover from that noise by inserting, deleting, or substituting characters while generating plaintext. Sequence-to-sequence models seem to be good candidates for this task.
\end{itemize}

\begin{figure}	
\captionsetup[subfigure]{labelformat=brace}
\begin{subfigure}{0.5\textwidth}		
	\includegraphics[scale=0.545]{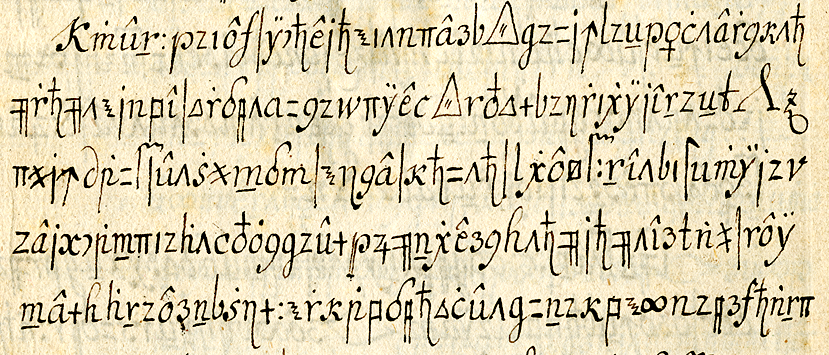}
	\caption{The Copiale cipher.\footnotemark} 	
	\label{fig:copiale}		
\end{subfigure}

\begin{subfigure}{0.5\textwidth}		
	\includegraphics[scale=0.63]{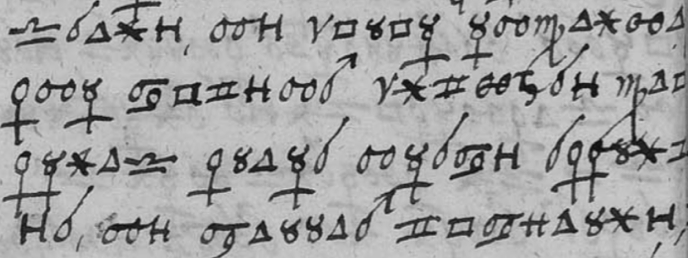}
	\caption{The Borg cipher.} 	
	\label{fig:borg}		
\end{subfigure}
	\caption{Historical cipher examples.} 	
	\label{fig:historical-ciphers}		
\end{figure}

\addtocounter{footnote}{-1}
 \stepcounter{footnote}\footnotetext{\url{https://cl.lingfil.uu.se/~bea/copiale/}}

\subsection{Decipherment as a Sequence-to-Sequence Translation Problem}
\label{sec:s2s-translation-decipherment}

\begin{figure*}	
\begin{subfigure}{1\textwidth}		
	\centering
	\includegraphics[scale=0.63]{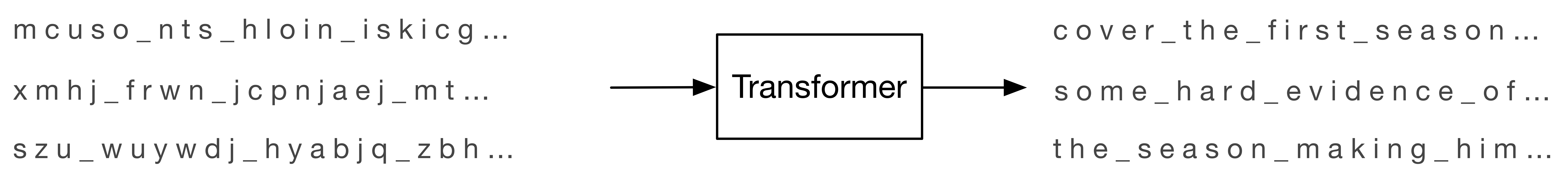}
	\caption{Input: Example ciphers encoded in random keys. Output: Plaintext in target language.} 	
	\label{fig:s2s-random}		
\end{subfigure}

\begin{subfigure}{1\textwidth}		
	\centering
	\includegraphics[scale=0.63]{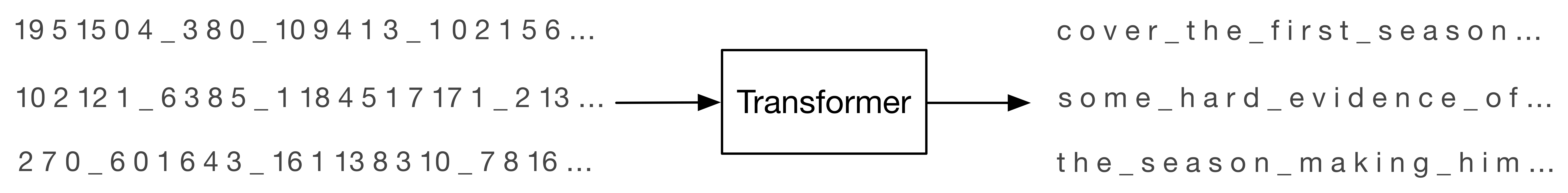}
	\caption{Input: Example ciphers encoded according to frequency ranks in descending order. Output: Plaintext in target language.} 	
	\label{fig:s2s-freq}		
\end{subfigure}
	\caption{Decipherment as a sequence-to-sequence translation problem. (\subref{fig:s2s-random}) shows the original ciphers being fed to the model. (\subref{fig:s2s-freq}) shows the same ciphers after frequency encoding.} 	

\end{figure*}

To cast decipherment as a supervised translation task, we need training data, i.e. pairs of <$f_1^N$, $e_1^M$> to train on. We can create this data using randomly generated substitution keys (Figure~\ref{fig:s2s-random}). We can then train a character-based sequence-to-sequence decipherment model and evaluate it on held-out text which is also encrypted with (different) randomly generated substitution keys. However, if we attempt this experiment using the Transformer model described in Section~\ref{sec:transformer}, we get abysmal results (see Section~\ref{sec:cipher-length} for scoring details).

Increasing the amount of training data won't help; there are $26! \approx 4 \times 10^{26}$ possible keys for English ciphers, and even if every key is represented, most of the training data will still be encoded with keys that are not used to encode the test data.  In fact, since each training \textit{example} uses a different key, we cannot assume that a character type has any particular meaning. The fundamental assumption behind embeddings is therefore broken. In the next section, we describe one way to overcome these challenges.

\subsection{Frequency Analysis}\label{sec:frequency-analysis}

To address the aforementioned challenges, we employ a commonly used technique in cryptanalysis called \textbf{frequency analysis}. Frequency analysis is attributed to the great polymath, Al-Kindi (801-873 C.E.) \citep{dooley-2013}. This technique has been used in previous decipherment work \cite{hauer-2016, kambhatla-2018}. It is based on the fact that in a given text, letters and letter combinations (n-grams) appear in varying frequencies, and that the character frequency distribution is \textit{roughly} preserved in any sample drawn from a given language. So, in different pairs of <$f_1^N$, $e_1^M$>, we expect the frequency distribution of characters to be similar. 

To encode that information, we re-map each ciphertext character to a value based on its frequency rank (Figure \ref{fig:s2s-freq}). This way, we convert any ciphertext to a ``frequency-encoded'' cipher. Intuitively, by frequency encoding, we are reducing the number of possible substitution keys (assuming frequency rank is roughly preserved across all ciphers from a given language). This is only an approximation, but it helps restore the assumption that there is a coherent connection between a symbol and its type embedding. For example, if the letters ``e'' and ``i'' are the most frequent characters in English, then in any 1:1 substitution cipher, they will be encoded as ``0'' or ``1'' instead of a randomly chosen character.

\subsection{The Transformer}
\label{sec:transformer}

We follow the character-based NMT approach in \citet{gao-2020} and use the Transformer model \cite{vaswani-2017} for our decipherment problem. The Transformer is an attention-based encoder-decoder model that has been widely used in the NLP community to achieve state-of-the-art performance on many sequence modeling tasks. We use the standard Transformer architecture, which consists of six encoder layers and six decoder layers as described in \citet{gao-2020}. 

\section{Data}
\label{sec:data}

For training, we create 1:1 substitution ciphers for 14 languages using random keys. For English, we use English Gigaword \cite{gigaword-2011}. We scrape historical text from Project Gutenberg for 13 other languages, namely: Catalan, Danish, Dutch, Finnish, French, German, Hungarian, Italian, Latin, Norwegian, Portuguese, Spanish, and Swedish.\footnote{Our dataset is available at \url{https://github.com/NadaAldarrab/s2s-decipherment}} Table \ref{tab:hist-data} summarizes our datasets. Following previous literature \cite{nuhn-2013, aldarrab-2017, kambhatla-2018}, we lowercase all characters and remove all non-alphabetic and non-space symbols. We make sure ciphers do not end in the middle of a word. We strip accents for languages other than English.

\begin{table}
	\centering
	\begin{tabular}{lrr}	
\hline
Language & Words & Characters \\ \hline
Catalan &  915,595  &  4,953,516  \\
Danish &  2,077,929  &  11,205,300  \\
Dutch &  30,350,145  &  177,835,527  \\
Finnish &  22,784,172  &  168,886,663  \\
French &  39,400,587  &  226,310,827  \\
German &  3,273,602  &  20,927,065  \\
Hungarian &  497,402  &  3,145,451  \\
Italian &  4,587,027  &  27,786,754  \\
Latin &  1,375,804  &  8,740,808  \\
Norwegian &  706,435  &  3,673,895  \\
Portuguese &  10,841,171  &  62,735,255  \\
Spanish &  20,165,731  &  114,663,957  \\
Swedish &  3,008,680  &  16,993,146  \\ \hline

	\end{tabular}
	\caption[Summary of datasets obtained from Project Gutenberg]{Summary of data sets obtained from Project Gutenberg.}
	\label{tab:hist-data}

\end{table}

\section{Experimental Evaluation}

To make our experiments comparable to previous work \cite{nuhn-2013, kambhatla-2018}, we create test ciphers from the English Wikipedia article about History.\footnote{\url{https://en.wikipedia.org/wiki/History}} We use this text to create ciphers of length 16, 32, 64, 128, and 256 characters. We generate 50 ciphers for each length. We follow the same pre-processing steps to create training data.

We carry out four sets of experiments to study the effect of cipher length, space encipherment/removal, unknown plaintext language, and transcription noise. Finally, we test our models on a real historical cipher, whose plaintext language was not known until recently.

As an evaluation metric, we follow previous literature \cite{kambhatla-2018} and use Symbol Error Rate (SER). SER is the fraction of incorrect symbols in the deciphered text. For space restoration experiments (Section~\ref{sec:nos-ciphers}), we use Translation Edit Rate (TER) \cite{snover-2006}, but on the character level. We define character-level TER as:

\begin{equation}
\mbox{TER} = \frac{\mbox{\# of edits}}{\mbox{\# of reference characters}}
\end{equation}

\noindent where possible edits include the insertion, deletion, and substitution of single characters. When the ciphertext and plaintext have equal lengths, SER is equal to TER.

We use FAIRSEQ to train our models \cite{fairseq-2019}. We mostly use the same hyperparameters as \citet{gao-2020} for character NMT, except that we set the maximum batch size to 10K tokens and use half precision floating point computation for faster training. The model has about 44M parameters. Training on a Tesla V100 GPU takes about 110 minutes per epoch. We train for 20 epochs. Decoding takes about 400 character tokens/s. We use a beam size of 100. Unless otherwise stated, we use 2M example ciphers to train, 3K ciphers for tuning, and 50 ciphers for testing in all experiments. We report the average SER on the 50 test ciphers of each experiment.

\subsection{Cipher Length}
\label{sec:cipher-length}

We first experiment with ciphers of length 256 using the approach described in Section~\ref{sec:s2s-translation-decipherment} (i.e. we train a Transformer model on pairs of <$f_1^N$, $e_1^M$> without frequency encoding). As expected, the model is not able to crack the 50 test ciphers, resulting in an SER of 71.75\%. For the rest of the experiments in this paper, we use the frequency encoding method described in Section~\ref{sec:frequency-analysis}.

Short ciphers are more challenging than longer ones. Following previous literature, we report results on different cipher lengths using our method. Table \ref{tab:cipher-length} shows decipherment results on ciphers of length 16, 32, 64, 128, and 256. For the 256 length ciphers, we use the aforementioned 2M train and 3K development splits. For ciphers shorter than 256 characters, we increase the number of examples such that the total number of characters remains nearly constant, at about 512M characters. We experiment with training five different models (one for each length) and training a single model on ciphers of mixed lengths. In the latter case, we also use approx. 512M characters, divided equally among different lengths. The results in Table~\ref{tab:cipher-length} show that our model achieves comparable results to the state-of-the-art model of \citet{kambhatla-2018} on longer ciphers, including perfect decipherment for ciphers of length 256. The table also shows that our method is more accurate than~\citet{kambhatla-2018} for shorter, more difficult ciphers of lengths 16 and 32. In addition, our method provides the ability to train on multilingual data, which we use to attack ciphers with an unknown plaintext language as described in Section~\ref{sec:multilingual-model}.

\begin{table*}
\centering
\begin{tabular}{llllll}
\hline
& \multicolumn{5}{c}{\textbf{Cipher Length}} \\
& \textbf{16} & \textbf{32} & \textbf{64} & \textbf{128} & \textbf{256} \\
\hline
Beam NLM \\ \citep{kambhatla-2018} & 26.80 & 5.80 & 0.07 & 0.01 & 0.00 \\
Beam (NLM + FreqMatch) \\ \citep{kambhatla-2018} & 31.00 & 2.90 & 0.07 & 0.02 & 0.00 \\
\hline
Transformer + Freq + separate models (this work) & 20.62 & \textbf{1.44} & 0.41 & 0.02 & 0.00 \\
Transformer + Freq + single model (this work) & \textbf{19.38} & 2.44 & 1.22 & 0.02 & 0.00 \\

\hline
\end{tabular}
\caption{\label{tab:cipher-length}
SER (\%) for solving 1:1 substitution ciphers of various lengths using our decipherment method.
}
\end{table*}

\subsection{No-Space Ciphers}
\label{sec:nos-ciphers}

The inclusion of white space between words makes decipherment easier because word boundaries can give a strong clue to the cryptanalyst. In many historical ciphers, however, spaces are hidden. For example, in the Copiale cipher (Figure \ref{fig:copiale}), spaces are enciphered with special symbols just like other alphabetic characters \cite{knight-etal-2011-copiale}. In other ciphers, spaces might be omitted from the plain text before enciphering, as was done in the Zodiac-408 cipher \cite{nuhn-2013}. We test our method in four scenarios:
\begin{enumerate}
    \item Ciphers with spaces (comparable to \citet{kambhatla-2018}).
    \item Ciphers with enciphered spaces. In this case, we treat space like other cipher characters during frequency encoding as described in Section~\ref{sec:frequency-analysis}.
    \item No-space ciphers. We omit spaces in both (source and target) sides.
    \item No-space ciphers with space recovery. We omit spaces from source but keep them on the target side. The goal here is to train the model to restore spaces along with the decipherment. \label{item:spacerecovery}
\end{enumerate}

Table~\ref{tab:nos-ciphers} shows results for each of the four scenarios on ciphers of length 256. During decoding, we force the model to generate tokens to match source length. Results show that the method is robust to both enciphered and omitted spaces. In scenario~\ref{item:spacerecovery}, where the model is expected to generate spaces and thus the output length differs from the input length, we limit the output to exactly 256 characters, but we allow the model freedom to insert spaces where it sees fit. The model generates spaces in accurate positions overall, leading to a TER of 1.88\%.

\begin{table}
\centering
\begin{tabular}{lr}
\hline
 \textbf{Cipher Type} & \textbf{TER(\%)} \\
\hline
Ciphers with spaces & 0.00 \\
Ciphers with enciphered spaces & 0.00 \\
No-space ciphers & 0.77 \\
No-space ciphers + generate spaces & 1.88 \\
\hline
\end{tabular}
\caption{\label{tab:nos-ciphers}
TER (\%) for solving 1:1 substitution ciphers of length 256 with different spacing conditions.}
\end{table}

\subsection{Unknown Plaintext Language}
\label{sec:multilingual-model}

While combing through libraries and archives, researchers have found many ciphers that are not accompanied with any cleartext or keys, leaving the plaintext language of the cipher unknown \cite{megyesi-2020}. To solve that problem, we train a single multilingual model on the 14 different languages described in Section~\ref{sec:data}. We train on a total of 2.1M random ciphers of length 256 (divided equally among all languages). We report results as the number of training languages increases while keeping the total number of 2.1M training examples fixed (Table~\ref{tab:multilingual-ciphers}). Increasing the number of languages negatively affects performance, as we expected. However, our experiments show that the 14-language model is still able to decipher 700 total test ciphers with an average SER of 0.68\%. Since we are testing on 256-character ciphers, this translates to no more than two errors per cipher on average.

\begin{table*}
\centering
\resizebox{\linewidth}{!}{
\begin{tabular}{l|rrrrrrrrrrrrrr|r}
\hline
 \textbf{\# lang} & \textbf{ca} & \textbf{da} & \textbf{nl} & \textbf{en}& \textbf{fi}& \textbf{fr} & \textbf{de} & \textbf{hu} & \textbf{it} & \textbf{la} & \textbf{no} & \textbf{pt}  & \textbf{es} & \textbf{sv} & \textbf{avg} \\
\hline
3 & - & - & - & 0.04 & - & 0.23 & - & - & - & - & - & - & 0.39 & - & 0.29 \\
7 & - & - & - & 0.08 & - & 0.34 & 0.30 & - & 1.23 & 1.38 & - & 0.48 & 0.40 & - & 0.60 \\
14 & 0.34 & 1.29 & 0.79 & 0.25 & 0.20 &  0.20 & 0.41 & 0.64 & 1.52 & 1.43 & 0.41 & 0.69 & 0.72 & 0.70 & 0.68\\
\hline
\end{tabular}
}
\caption{\label{tab:multilingual-ciphers}
SER (\%) for solving 1:1 substitution ciphers using a multilingual model trained on a different number of languages. Each language is evaluated on 50 test ciphers generated with random keys.
}
\end{table*}

\subsection{Transcription Noise}
\label{sec:noise}

Real historical ciphers can have a lot of noise. This noise can come from the natural degradation of historical documents, human mistakes during a manual transcription process, or misspelled words by the author, as in the Zodiac-408 cipher. Noise can also come from  automatically transcribing historical ciphers using Optical Character Recognition (OCR) techniques \cite{yin-19}. It is thus crucial to have a robust decipherment model that can still crack ciphers despite the noise.

\citet{hauer-2014} test their proposed method on noisy ciphers created by randomly corrupting $log_{2}(N)$ of the ciphertext characters. However, automatic transcription of historical documents is very challenging and can introduce more types of noise, including the addition and deletion of some characters during character segmentation \cite{yin-19}. We test our model on three types of random noise: insertion, deletion, and substitution. We experiment with different noise percentages for ciphers of length 256 (Table~\ref{tab:noisy-ciphers}). We report the results of training (and testing) on ciphers with only substitution noise and ciphers that have all three types of noise (divided equally). We experimentally find that training the models with 10\% noise gives the best overall accuracy, and we use those models to get the results in Table~\ref{tab:noisy-ciphers}. Our method is able to decipher with up to 84\% accuracy on ciphers with 20\% of random insertion, deletion, and substitution noise. Figure \ref{fig:noise-examples} shows an example output for a cipher with 15\% noise. The model recovers most of the errors, resulting in a TER of 5.86\%. One of the most challenging noise scenarios, for example, is the deletion of the last two characters from the word ``its.'' The model output the word ``i,'' which is a valid English word. Of course, the more noise there is, the harder it is for the model to recover due to error accumulation.

\begin{table}
\centering
\begin{tabular}{crr}
\hline
& \multicolumn{2}{c}{\textbf{Noise Type}} \\
\hline
 \textbf{\% Noise} & \textbf{sub} & \textbf{sub, ins, del} \\
\hline
5 & 1.10 & 2.87 \\
10 & 2.40 & 5.87 \\
15 & 5.28 & 10.58 \\
20 & 11.48 & 16.17 \\
25 & 17.63 & 27.43 \\
\hline
\end{tabular}
\caption{\label{tab:noisy-ciphers}
TER (\%) for solving 1:1 substitution ciphers with random insertion, deletion, and substitution noise. These models have been trained with 10\% noise.
}
\end{table}

\begin{figure*}[ht]
\centering
\begin{tabular}{p{0.07\linewidth} | p{0.88\linewidth}}
\hline
Source &  {{\fontsize{11}{13.2}\selectfont 3 2 11 11 2 6 4 15 0 \_ 16 0 1 6 \_ \textcolor{red}{d} 20 12 9 \textcolor{red}{i5} 2 4 3 1 \_ 2 3 \_ \textcolor{red}{d} 15 0 3 6 \_ 2 \textcolor{red}{s22} \_ 18 \textcolor{red}{i16} 0 9 9 \_ 2 1 \_ 6 13 0 \_ 1 4 \textcolor{red}{i7} 19 3 4 5 4 10 2 3 \textcolor{red}{i13} 10 0 \_ 7 5 \_ 8 \textcolor{red}{d} 5 5 0 11 0 3 6 \_ 10 2 14 1 0 \textcolor{red}{i21} 1 \_ 2 3 8 \_ 0 5 5 0 10 6 1 \colorbox{yellow}{\textcolor{red}{i0}} \_ 13 4 1 6 7 \textcolor{red}{s5} 4 2 3 \textcolor{red}{s6} \_ 2 9 1 7 \textcolor{red}{i18} \_ 8 0 16 2 6 0 \_ 6 13 0 \_ 3 2 6 14 \textcolor{red}{d} 0 \_ \textcolor{red}{s3} 5 \_ \colorbox{yellow}{\textcolor{red}{d}} 4 1 6 7 \colorbox{yellow}{\textcolor{red}{d}} 17 \_ \textcolor{red}{s5} 3 8 \_ 4 \colorbox{yellow}{\textcolor{red}{d}} \colorbox{yellow}{\textcolor{red}{d}} \_ 14 1 0 5 \textcolor{red}{s0} 9 3 0 1 1 \_ 16 17 \_ 8 \textcolor{red}{i5} 4 1 10 14 1 1 4 \textcolor{red}{s23} 19 \_ \textcolor{red}{s2} 13 0 \_ 1 \textcolor{red}{s11} 14 \textcolor{red}{s24} 17 \_ 7 5 \_ 6 13 \textcolor{red}{i21} 0 \_ 8 4 1 10 4 12 9 4 3 0 \_ 2 1 \_ 2 \textcolor{red}{i7} 3 \_ 0 3 8 \_ 4 3 \_ \textcolor{red}{s5} 6 1 0 \textcolor{red}{s14} \textcolor{red}{s12} \_ 2 3 8 \_ 1 \textcolor{red}{d} \_ 2 \_ 18 \textcolor{red}{d} 17 \_ 7 \textcolor{red}{i20} 5 \textcolor{red}{i9} \_ 12 11 7 15 4 8 4 \textcolor{red}{s2} 19 \_ 12 0 11 \textcolor{red}{i12} 0 \textcolor{red}{d} \textcolor{red}{d} 10 \textcolor{red}{d} 4 15 0}}\\
\hline
Target & { n a r r a t i v e \_ b e s t \_ e x p l a i n s \_ a n \_ e v e n t \_ a s \_ w e l l \_ a s \_ t h e \_ s i g n i f i c a n c e \_ o f \_ d i f f e r e n t \_ c a u s e s \_ a n d \_ e f f e c t \colorbox{yellow}{s} \_ h i s t o r i a n s \_ a l s o \_ d e b a t e \_ t h e \_ n a t u r e \_ o f \_ \colorbox{yellow}{h} i s t o r \colorbox{yellow}{y} \_ a n d \_ i \colorbox{yellow}{t s} \_ u s e f u l n e s s \_ b y \_ d i s c u s s i n g \_ t h e \_ s t u d y \_ o f \_ t h e \_ d i s c i p l i n e \_ a s \_ a n \_ e n d \_ i n \_ i t s e l f \_ a n d \_ a s \_ a \_ w a y \_ o f \_ p r o v i d i n g \_ p e r s p e c t i v \colorbox{yellow}{e} }\\
\hline 
Output &  { n a r r a t i v e \_ b e s t \_ e x p l a i n s \_ a n \_ e v e n t \_ a s \_ w e l l \_ a s \_ t h e \_ s i g n i f i c a n c e \_ o f \_ d i f f e r e n t \_ c a u s e s \_ a n d \_ e f f e c t \colorbox{yellow}{i v e} \_ h i s t o r i a n s \_ a l s o \_ d e b a t e \_ t h e \_ n a t u r e \_ o f \_ \colorbox{yellow}{v} i s \colorbox{yellow}{i} t o r \colorbox{yellow}{s} \_ a n d \_ i \colorbox{yellow}{ } \_ u s e f u l n e s s \_ b y \_ d i s c u s s i n g \_ t h e \_ s t u d y \_ o f \_ t h e \_ d i s c i p l i n e \_ a s \_ a n \_ e n d \_ i n \_ i t s e l f \_ a n d \_ a s \_ a \_ w a y \_ o f \_ p r o v i d i n g \_ p e r s p e c t i v \colorbox{yellow}{ }}\\
\hline
\end{tabular}

\caption{Example system output for a cipher with 15\% random noise (shown in red). Substitutions, insertions, and deletions are denoted by letters s, i, and d, respectively. The system recovered 34/40 errors (TER is 5.86\%). Highlighted segments show the errors that the system failed to recover from.}
\label{fig:noise-examples}
\end{figure*}

\subsection{The Borg Cipher}
The Borg cipher is a 400-page book digitized by the Biblioteca Apostolica Vaticana (Figure \ref{fig:borg}).\footnote{\url{http://digi.vatlib.it/view/MSS_Borg.lat.898}.} The first page of the book is written in Arabic script, while the rest of the book is enciphered using astrological symbols. The Borg cipher was first automatically cracked by \citet{aldarrab-2017} using the noisy-channel framework described in \citet{knight-etal-2006-unsupervised}. The plaintext language of the book is Latin. The deciphered book reveals pharmacological knowledge and other information about that time.

\begin{figure}[H]	
\begin{tabular}{l}
\hline
\includegraphics[scale=0.35]{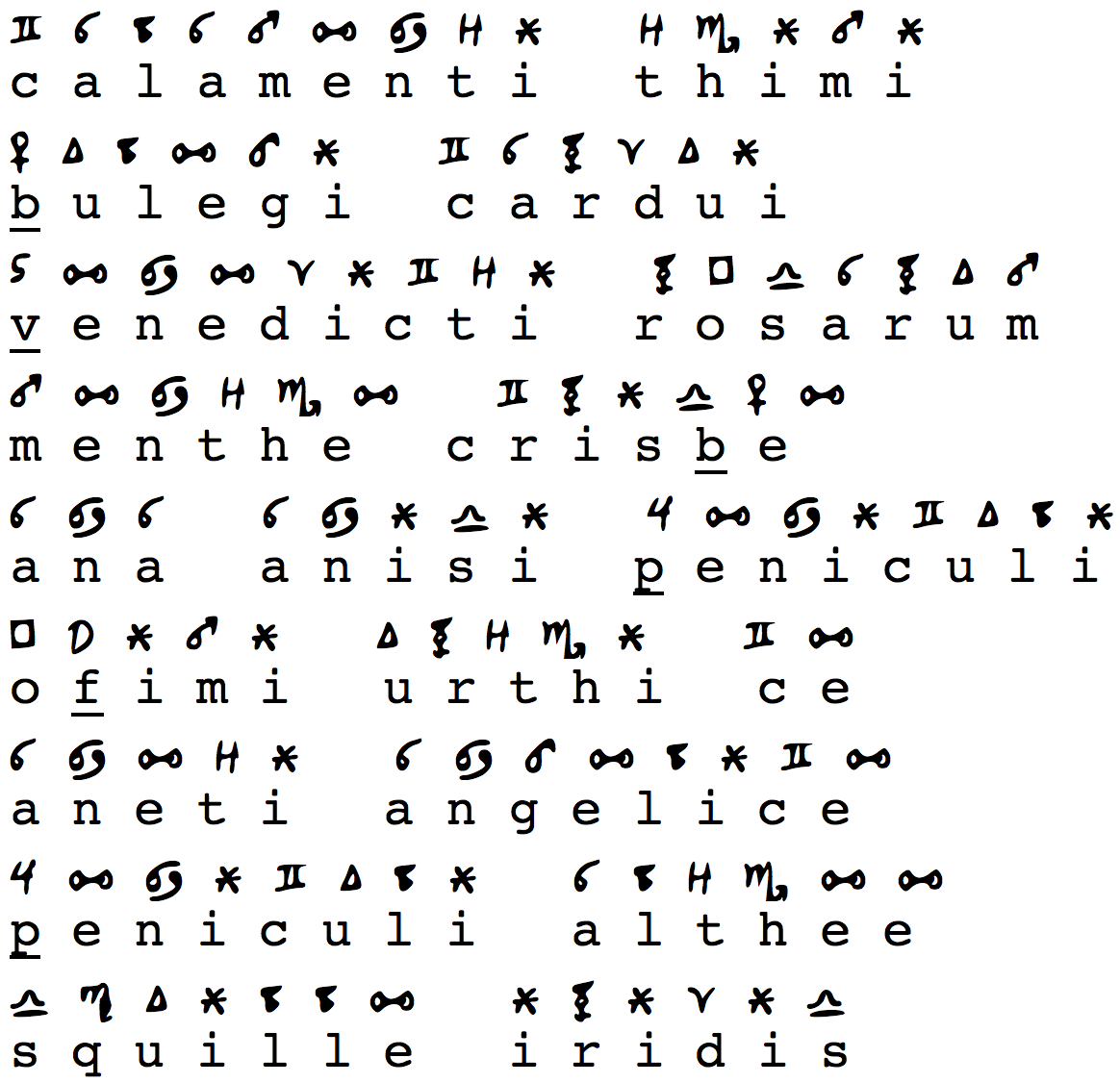}\\
\hline
\end{tabular}
	\caption{The first 132 characters of the Borg cipher and its decipherment. Errors are underlined. Correct words are: \underline{p}ulegi, \underline{b}enedicti, cris\underline{p}e, o\underline{z}imi, and \underline{f}eniculi.} 	
	\label{fig:borg-decipherment}		
\end{figure}

We train a Latin model on 1M ciphers and use the first 256 characters of the Borg cipher to test our model. Our model is able to decipher the text with an SER of 3.91\% (Figure~\ref{fig:borg-decipherment}). We also try our 14-language multilingual model on this cipher, and obtain an SER of 5.47\%. While we cannot directly compare to \citet{aldarrab-2017}, who do not report SER, this is a readable decipherment and can be easily corrected by Latin scholars who would be interested in such a text.

\section{Anagram Decryption}

To further test the capacity of our model, we experiment with a special type of noise. In this section, we address the challenging problem of solving substitution ciphers in which letters within each word have been randomly shuffled. Anagramming is a technique that can be used to further disguise substitution ciphers by permuting characters. Various theories about the mysterious Voynich Manuscript, for example, suggest that some anagramming scheme was used to encode the manuscript \cite{reddy-knight-2011-know}. \citet{hauer-2016} propose a two-step approach to solve this problem. First, they use their 1:1 substitution cipher solver \cite{hauer-2014} to decipher the text. The solver is based on tree search for the key, guided by character-level and word-level n-gram language models. They adapt the solver by relaxing the letter order constraint in the key mutation component of the solver. They then re-arrange the resulting deciphered characters using a word trigram language model. 

We try a one-step, end-to-end anagram decryption model. In our sequence-to-sequence formulation, randomly shuffled characters can confuse the training. We thus represent an input cipher as a bag of frequency-mapped characters, nominally presented in frequency rank order (Figure \ref{fig:anagram}). We use the English Gigaword dataset to train a 256 character model on the sorted frequencies and test on the aforementioned test set of 50 ciphers (after applying random anagramming). Following \citet{hauer-2016}, we report word accuracy on this task. Our model achieves a word accuracy of 95.82\% on the 50 Wikipedia ciphers. 

\citet{hauer-2016} report results on a test set of 10 long ciphers extracted from 10 Wikipedia articles about art, Earth, Europe, film, history, language, music, science, technology, and Wikipedia. Ciphers have an average length of 522 characters. They use English Europarl to train their language models \cite{koehn2005epc}. To get comparable results, we trained a model on ciphers of length 525 created from the English side of the Spanish-English Europarl dataset. Our model achieved a word accuracy of 96.05\% on \citeauthor{hauer-2016}'s test set. Training on English Gigaword gave a word accuracy of 97.16\%, comparable to the 97.72\% word accuracy reported by \citet{hauer-2016}. This shows that our simple model can crack randomly anagrammed ciphers, which hopefully inspires future work on other cipher types.

\begin{figure*}
(1) \texttt{t h e \_ i n v e n t i o n \_ o f \_ w r i t i n g \_ s y s t e m s}\\
(2) \texttt{j c z \_ m r b z r j m k r \_ k f \_ w u m j m r e \_ a o a j z g a} \\
(3) \texttt{c j z \_ k z m r b r j m r \_ f k \_ e w u j m m r \_ z g o a j a a} \\
(4) \texttt{6 0 3 \_ 5 3 1 2 7 2 0 1 2 \_ 8 5 \_ 11 9 10 0 1 1 2 \_ 3 13 12 4 0 4 4}\\
(5) \texttt{0 3 6 \_ 0 1 1 2 2 2 3 5 7 \_ 5 8 \_ 0 1 1 2 9 10 11 \_ 0 3 4 4 4 12 13}\\
(6) \texttt{t h e \_ i n v e n t i o n \_ o f \_ \colorbox{yellow}{b r i t a i n} \_ s y s t e m s}
\caption{Example anagram encryption and decryption process: (1) original plaintext (2) after applying a 1:1 substitution key (3) after anagramming (this is the ciphertext) (4) after frequency encoding (5) after sorting frequencies. This is fed to Transformer (6) system output (errors are highlighted).}
\label{fig:anagram}
\end{figure*}

\section{Related Work}

Deciphering substitution ciphers is a well-studied problem in the natural language processing community, e.g., \cite{hart-94, olsun-07, ravi-2008, corlett-2010, nuhn-2013, nuhn-2014, hauer-2014, aldarrab-2017}. Many of the recent proposed methods search for the substitution table (i.e. cipher key) that leads to a likely target plaintext according to a character n-gram language model. The current state-of-the-art method uses beam search and a neural language model to score candidate plaintext hypotheses from the search space for each cipher, along with a frequency matching heuristic incorporated into the scoring function \cite{kambhatla-2018}. This method, which is comparable in results to our method on longer ciphers and slightly weaker on shorter ciphers, assumes prior knowledge of the target plaintext language. Our method, by contrast, can solve substitution ciphers from different languages without explicit language identification.

Recent research has looked at applying other neural models to different decipherment problems. \citet{Greydanus-17} find an LSTM model can learn the decryption function of polyalphabetic substitution ciphers when trained on a concatenation of <key + ciphertext> as input and plaintext as output. Our work looks at a different problem. We target a ciphertext-only-attack for short 1:1 substitution ciphers. \citet{gomez-2018} propose CipherGAN, which uses a Generative Adversarial Network to find a mapping between the character embedding distributions of plaintext and ciphertext. This method assumes the availability of plenty of ciphertext. Our method, by contrast, does not require a large amount of ciphertext. In fact, all of our experiments were evaluated on ciphers of 256 characters or shorter.

Early work on language identification from ciphertext uses the noisy-channel decipherment model \cite{knight-etal-2006-unsupervised}. Specifically, the expectation-maximization algorithm is used to learn mapping probabilities, guided by a pre-trained n-gram language model. This decipherment process is repeated for all candidate languages. The resulting decipherments are ranked based on the probability of the ciphertext using the learned model, requiring a brute-force guess-and-check approach that does not scale well as more languages are considered. \citet{hauer-2016} use techniques similar to ours, incorporating  character frequency, decomposition pattern frequency, and trial decipherment in order to determine  the language of a ciphertext.

\section{Conclusion and Future Work}

In this work, we present an end-to-end decipherment model that is capable of solving simple substitution ciphers without the need for explicit language identification. We use frequency analysis to make it possible to train a multilingual Transformer model for decipherment. Our method is able to decipher 700 ciphers from 14 different languages with less than 1\% SER. We apply our method on the Borg cipher and achieve 5.47\% SER using the multilingual model and 3.91\% SER using a monolingual Latin model. In addition, our experiments show that these models are robust to different types of noise, and can even recover from many of them. To the best of our knowledge, this is the first application of sequence-to-sequence neural models for decipherment.

We hope that this work drives more research in the application of contextual neural models to the decipherment problem. It would be interesting to develop other techniques for solving more complex ciphers, e.g.~homophonic and polyalphabetic ciphers.

\section*{Acknowledgements}

This research is based upon work supported by the Office of the Director of National Intelligence (ODNI), Intelligence Advanced Research Projects Activity (IARPA), via AFRL Contract FA8650-17-C-9116.  The views and conclusions contained herein are those of the authors and should not be interpreted as necessarily representing the official policies or endorsements, either expressed or implied, of the ODNI, IARPA, or the U.S. Government. The U.S. Government is authorized to reproduce and distribute reprints for Governmental purposes notwithstanding any copyright annotation thereon.

\section*{Ethics Statement}
 
This work, like all decipherment work, is concerned with the decoding of encrypted communications, and thus the methods it describes are designed to reveal information that has been deliberately obfuscated and thus violate the privacy of the authors. However, the class of problems it addresses, 1:1 substitution ciphers, are known to be relatively weak forms of encryption, once popular, but long considered obsolete. Thus, the major practical use of this work as a decryption tool is in the ability to quickly decode ancient ciphertexts, such as the Borg cipher, the contents of which are interesting for historical purposes but are not in danger of revealing secrets of any living person. Modern encryption schemes such as RSA, Blowfish, or AES cannot be defeated by the methods presented here.

We have demonstrated our work's effectiveness on ciphers of 14 alphabetic languages. The approaches presented here may be less effective on other orthographic systems such as abjads (which have fewer explicit symbols and more inherent ambiguity), abugidas (which have more explicit symbols and thus are conceivably less tractable), or logographic systems (which have many more explicit symbols). We caution that more exploration needs to be done before relying on the methods presented here when decoding ancient historical ciphertexts that are not encodings of alphabetic plaintext.

It is possible, though unlikely, that incorrect conclusions can be drawn if the approaches presented in this work yield false results. For instance, in Figure~\ref{fig:borg}, the word decoded as \texttt{peniculi} (towels) should in fact be decoded as \texttt{feniculi} (fennel); similar examples can be seen in Figure~\ref{fig:noise-examples}. The translation ``seed of towels'' being far less likely than ``seed of fennel`` in context, we would expect easy detection of this kind of error. We recommend that these methods not be trusted exclusively, but rather that they be used as one tool in a cryptologist's kit, alongside language expertise and common sense, such that incoherent decodings may be given a careful look and correction.

\bibliographystyle{acl_natbib}
\bibliography{acl2021}

\end{document}